\documentclass[lettersize,journal]{IEEEtran}
\usepackage{amsmath,amsfonts}
\usepackage{algorithmic}
\usepackage{algorithm}
\usepackage{array}
\usepackage[caption=false,font=normalsize,labelfont=sf,textfont=sf]{subfig}
\usepackage{textcomp}
\usepackage{stfloats}
\usepackage{url}
\usepackage{verbatim}
\usepackage{graphicx}
\usepackage{cite}
\usepackage{algorithm}  
\usepackage[ruled,vlined,algo2e]{algorithm2e}  
\usepackage{pifont}  
\usepackage{booktabs}
\usepackage{colortbl} 
\usepackage{makecell}
\usepackage{array}
\usepackage{multirow} 
\usepackage{tabularx}
\usepackage{hyperref}
\usepackage{graphicx} 
\usepackage{amssymb}
\usepackage{xcolor}
\begin{document}

\title{Efficient Non-Exemplar Class-Incremental Learning with Retrospective Feature Synthesis}

\author{Liang Bai, Hong Song, Yucong Lin, Tianyu Fu, Deqiang Xiao, Danni Ai, Jingfan Fan, Jian Yang
\thanks{
Corresponding authors: Hong Song and Jian Yang.
}

\thanks{Liang Bai and Hong Song are with the School of Computer Science and Technology, Beijing Institute of Technology, Beijing 100081, China.}

\thanks{ Yucong Lin, Deqiang Xiao, Danni Ai, Jingfan Fan, and Jian Yang are with the School of Optics and Photonics, Beijing Institute of Technology, Beijing 100081, China.}

\thanks{Tianyu Fu is with the School of Medical Technology, Beijing Institute of Technology, Beijing 100081, China.}
}

\markboth{Journal of \LaTeX\ Class Files,~Vol.~14, No.~8, August~2021}%
{Shell \MakeLowercase{\textit{et al.}}: A Sample Article Using IEEEtran.cls for IEEE Journals}


\maketitle

\begin{abstract}
Despite the outstanding performance in many individual tasks, deep neural networks suffer from catastrophic forgetting when learning from continuous data streams in real-world scenarios. Current Non-Exemplar Class-Incremental Learning (NECIL) methods mitigate forgetting by storing a single prototype per class, which serves to inject previous information when sequentially learning new classes. However, these stored prototypes or their augmented variants often fail to simultaneously capture spatial distribution diversity and precision needed for representing old classes. Moreover, as the model acquires new knowledge, these prototypes gradually become outdated, making them less effective. To overcome these limitations, we propose a more efficient NECIL method that replaces prototypes with synthesized retrospective features for old classes. Specifically, we model each old class's feature space using a multivariate Gaussian distribution and generate deep representations by sampling from high-likelihood regions. Additionally, we introduce a similarity-based feature compensation mechanism that integrates generated old class features with similar new class features to synthesize robust retrospective representations. These retrospective features are then incorporated into our incremental learning framework to preserve the decision boundaries of previous classes while learning new ones. Extensive experiments on CIFAR-100, TinyImageNet, and ImageNet-Subset demonstrate that our method significantly improves the efficiency of non-exemplar class-incremental learning and achieves state-of-the-art performance.

\end{abstract}


\section{Introduction}

\IEEEPARstart
{A}{utomatic} visual recognition based on deep neural networks is a fundamental task in computer vision that has undergone substantial advancements in recent years \cite{ref_fine_cls_1,ref_tcs_cls,ref_fine_cls_2}. However, most existing methods assume a static scenario, where the entire training process occurs in a single session with all training data fully available. In contrast, real-world scenarios are dynamic, with new data continuously emerging, requiring continuous updates to the model. Therefore, it is essential to develop efficient visual recognition algorithms that can learn continuously, much like the human ability to acquire new concepts without forgetting previously learned knowledge.

\begin{figure}[tbp]
\centering
\captionsetup[subfigure]{font=footnotesize, labelfont={footnotesize}} 
\subfloat[\scriptsize Baseline]{\includegraphics[width=0.23\textwidth]{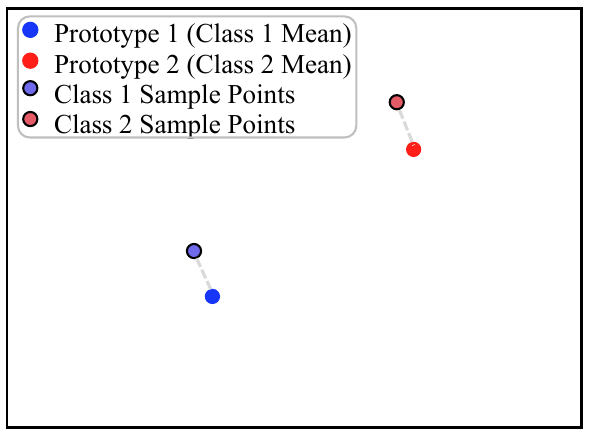}\label{fig:chart_a}}
\hfil
\subfloat[\scriptsize Gaussian Noise Aug]{\includegraphics[width=0.23\textwidth]{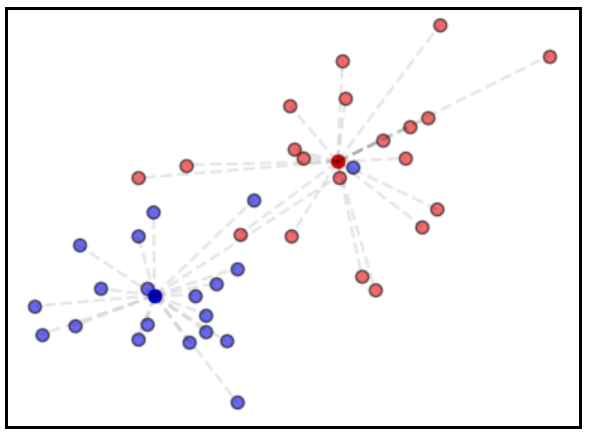}\label{fig:chart_b}}\\ \vspace{-0.5em}
\subfloat[\scriptsize Prototype Mixing]{\includegraphics[width=0.23\textwidth]{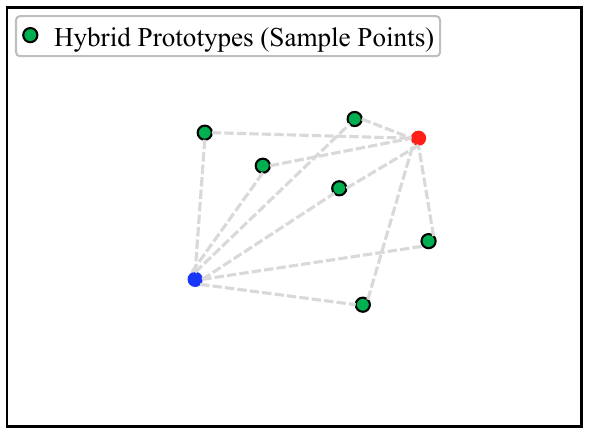}\label{fig:chart_c}}
\hfil
\subfloat[\scriptsize Our Method]{\includegraphics[width=0.23\textwidth]{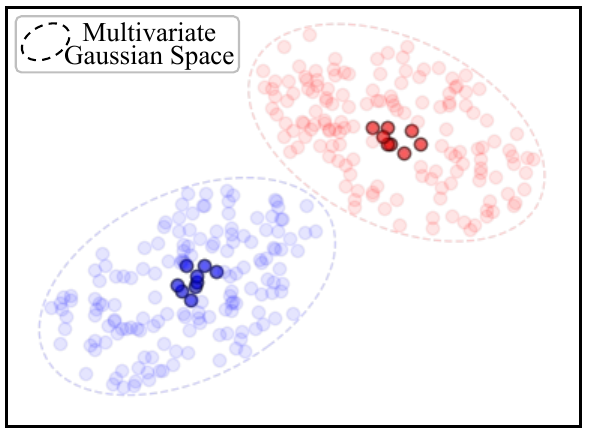}\label{fig:chart_d}}\vspace{-0.3em}
\caption{\small Idea illustration. (a) Baseline: Use the stored prototypes as the classification features of the old classes to rebuild the classification boundaries. (b) Gaussian Noise Aug \cite{ref_pass}: Use random Gaussian noise to enhance the prototypes, but it introduces inaccurate features and overlaps between class distributions due to noise interference to the representation space. (c) Prototype Mixing \cite{ref_rrfe}: Combine the prototypes of the old classes randomly linearly to form the hybrid prototypes, but these obtained prototypes lack authenticity. (d) Our Method: Model the representation space of each old class using a multivariate Gaussian distribution, then generate classification features for old classes by sampling from the high-likelihood regions, thereby better capturing the diversity and accuracy of the generated representations.}\vspace{-12pt}
\label{fig:chart}
\end{figure}

Class incremental learning \cite{ref_cil_sruvey,ref_tcs_cil1,ref_tcs_cil2} is a real-world scenario that considers adding new classes to an already trained classification model. This approach aims to mitigate the catastrophic forgetting that neural networks underfit previously learned classes while absorbing new class data \cite{ref_fetril}. Catastrophic forgetting typically arises due to the overlap or confusion between old and new class representations within the feature space \cite{ref_pass}. As a large number of new class instances are introduced,  the unified classifier is biased to these new classes,  causing the decision boundaries of previous classes to shift significantly\cite{ref_pass}. Some methods \cite{ref_icarl,ref_replay_gss,ref_replay_gem,ref_replay_rebalance,ref_repaly_foster} mitigate this issue by replaying a limited set of exemplars from old classes while learning new classes. However, past data is often unavailable due to privacy concerns or restrictions on the long-term storage of raw samples in real-world applications. Therefore, in this paper, we focus on Non-Exemplar Class-Incremental Learning (NECIL), a more challenging scenario where no raw data from previous tasks is retained.

In NECIL research, one of the primary challenges is to reconstruct the decision boundaries of previous classes without accessing their original samples. Some methods \cite{ref_generative_replay_1,ref_generative_replay_2,ref_generative_replay_3,ref_tcs_replay} leverage deep generative models to generate pseudo-samples of previous classes. However, these methods frequently encounter issues like model collapse and unreliable sample generation. To address these limitations, prototype-based methods have gained popularity for their ability to balance privacy security with robust representational capability \cite{ref_pass,ref_ssre,ref_praka,ref_rrfe,ref_fetril}. As shown in Figure \ref{fig:chart_a}, PASS \cite{ref_pass} first proposes storing a representative prototype for each old class, typically the class mean in the deep feature space. When learning new classes, these stored prototypes are used as old class representations to inject previous information. Furthermore, as illustrated in Figure \ref{fig:chart_b}, PASS \cite{ref_pass} augments these prototypes by adding Gaussian noise to increase the diversity of the injected representations. However, this method struggles with maintaining representation quality and ensuring clear distinctions between classes, as the added noise can interfere with the feature space. RRFE \cite{ref_rrfe}, depicted in Figure \ref{fig:chart_c}, randomly combines old class prototypes linearly to create hybrid prototypes with mixed labels. While innovative, these hybrid prototypes lack authenticity, making it difficult to accurately reconstruct the original class boundaries. Other methods, such as SSRE \cite{ref_ssre} and FeTrIL \cite{ref_fetril}, either directly use these augmented prototype variants or introduce enhancements based on them. Overall, these methods often overlook the distribution's shape and scope in the feature space of old classes, resulting in imprecise or limited diversity in class representations.

To overcome this challenge, we propose a Multivariate Gaussian Sampling (MGS) strategy, which offers a more robust generation of old class representations, replacing traditional prototypes. As depicted in Figure \ref{fig:chart_d}, our approach models the feature space of each old class using a multivariate Gaussian distribution. By sampling from the high-likelihood regions of these distributions, we generate diverse and high-quality representations of old classes. This strategy is inspired by recent work in out-of-distribution detection \cite{ref_vos}. Unlike prototype-based methods, which focus primarily on the class mean, our approach accounts for both the shape and scope of the entire feature distribution space. As a result, the generated old class representations exhibit both good diversity and higher accuracy and can more effectively reconstruct the decision boundaries of previous classes.

In addition to generating old class representations, another often overlooked challenge in NECIL is the increasing deviation between the classifier and the generated old class representations. Specifically, as the model continuously learns new class data, both the feature extractor and the classifier must be updated to maintain the model's plasticity. This ongoing process gradually leads to the outdated of the old class representations, whether obtained through prototype-based methods or our proposed strategy \cite{ref_praka}. In NECIL, this deviation is particularly problematic because it cannot be directly measured without access to raw samples from the old classes. To alleviate this issue, PRAKA \cite{ref_praka} introduces random bidirectional interpolation between new class features and stored old class prototypes to enhance the quality of old class representations. While this random interpolation somewhat mitigates representation deviation in scenarios with a limited number of incremental phases, it also introduces instability in the generated features due to its reliance on a random interpolation factor. As a result, representation deviation may increase over longer sequences of incremental tasks, leading to performance degradation, as demonstrated in the experimental results in Section \ref{sec_overall}.

To further address this bottleneck, we propose a Similarity-based Feature Compensation (SFC) mechanism to reduce the deviation between generated representations and the evolving classifier. Specifically, we use normalized cosine similarity to measure the similarity between the arriving new class features and generated old class features, selecting the most similar new class feature for each generated old class feature. We then apply a simple yet effective element-wise averaging between the generated and most similar feature sets to form the retrospective representations of old classes. These compensated features, along with their original labels, are used to retain the previously acquired knowledge. This compensatory mechanism ensures that synthesized old class representations not only remain closely aligned with real distributions over longer incremental tasks but also preserve a stable ability to reconstruct classification boundaries.

Finally, we build a robust baseline model that integrates self-supervised label augmentation \cite{ref_ssl_la} and knowledge distillation \cite{ref_kd} to combine the two key components described above, forming the proposed efficient non-exemplar class-incremental learning method with \textbf{R}etrospective \textbf{F}eature \textbf{S}ynthesis, termed RFS. Our main contributions are as follows:

\begin{itemize}
\item We propose an effective old class representation generation strategy for NECIL, which models the feature space with multivariate Gaussian distribution and sample representations from high-likelihood regions to resist catastrophic forgetting.

\item We contribute a similarity-based feature compensation mechanism, which selects the most similar new class features to compensate for the generated old class features through element-wise averaging, further improving the efficiency of incremental learning.

\item Extensive experiments and analysis demonstrate that our method achieves state-of-the-art performance on the NECIL benchmarks of CIFAR-100, TinyImageNet, and ImageNet-Subset.

\end{itemize}

\section{Related Work}

\subsection{Class Incremental Learning}
Class-Incremental Learning (CIL) addresses the challenge of continuous learning in real-world scenarios where new data classes are dynamically introduced. By providing flexible expansion capabilities to the existing end-to-end models, CIL has broad prospects and applications in object recognition \cite{ref_cil_survey_1,ref_cil_survey_2,ref_cil_survey_3}, object detection \cite{ref_fiod,ref_eiod,ref_meta_iod}, semantic segmentation \cite{ref_cil_seg_sur,ref_seg_med,ref_seg_med_2}, and other related fields. A key issue in CIL is mitigating the performance degradation on previously learned classes as models incrementally learn new ones. To tackle this, researchers have developed various CIL methods \cite{ref_cil_survey_1,ref_cil_survey_2,ref_cil_survey_3}, typically categorized into three types: parameter isolation methods, regularization-based methods, and replay-based methods. 

Parameter isolation methods assign different sets of parameters to different tasks, preventing interference between tasks during continuous learning \cite{ref_cil_survey_1}. Some approaches \cite{ref_cil_packnet,ref_cil_pathnet,ref_cil_hat} maintain a fixed model architecture, activating only task-specific parameters. Other works \cite{ref_cil_pnn,ref_cil_expert_gate,ref_cil_dan} employ dynamically growing network architectures, adding new parameters for new tasks while preserving those from previous tasks. While parameter isolation effectively avoids catastrophic forgetting, it limits knowledge sharing across tasks and can lead to rapid model growth or decreased accuracy, as sparse knowledge density hinders adaptability to more tasks.

Regularization-based methods introduce additional terms into the loss function to consolidate previously acquired knowledge while learning new tasks, avoiding the need to store raw samples. This approach prioritizes privacy security and reduces memory requirements \cite{ref_cil_survey_1}. For example, LwF \cite{ref_cil_lwf} uses the previous model’s output as a soft label to constrain the current model’s output, ensuring the current model behaves similarly to the previous one on old classes, thereby alleviating forgetting. EWC \cite{ref_ewc} selectively reduces plasticity in specific weights to protect previously learned knowledge. DMC \cite{ref_cil_dmc} trains a separate model for new classes, then uses double distillation to combine it with models trained on old classes, addressing performance limitations caused by asymmetric supervision between new and old classes.

Replay-based methods store raw samples or use generative models to create pseudo-samples of old classes, replaying these samples while learning new tasks to alleviate forgetting \cite{ref_cil_survey_1}. Methods like iCaRL \cite{ref_icarl}, ER \cite{ref_cil_er}, and SER \cite{ref_cil_ser} store a limited subset of raw samples to preserve recognition ability on old classes. However, these methods face challenges when raw samples cannot be stored due to privacy concerns or long-term data storage limitations. Some methods \cite{ref_generative_replay_1,ref_generative_replay_2,ref_generative_replay_3} generate pixel-level samples by modeling the pixel space of previous classes, but this increases the complexity of training the generative model and often results in issues such as model collapse and unreliable sample generation \cite{ref_cil_survey_1}. In this paper, we bypass the complexity and instability of pixel-level sample generation by directly generating high-dimensional features for old classes, offering higher efficiency and more reliable quality.

\subsection{Non-Exemplar Class-Incremental Learning}
In real-world applications, data privacy concerns and limitations on storing original data have increased interest in studying non-exemplar class-incremental learning methods. PASS \cite{ref_pass} stores a representative prototype for each old class and enhances these prototypes with Gaussian noise to maintain decision boundaries for previous tasks. SSRE \cite{ref_ssre} introduces a self-sustaining representation expansion scheme that preserves invariant knowledge while selectively incorporating new samples into the distillation process to improve discrimination between old and new classes. FeTrIL \cite{ref_fetril} combines a fixed feature extractor with a pseudo-feature generator, which creates old class representations by using the geometric translation of new class features, balancing stability and plasticity in incremental learning. PRAKA \cite{ref_praka} utilizes random bidirectional interpolation to blend previous class prototypes with new class features, dynamically reshaping old class feature distributions to preserve classification decision boundaries. Additionally, RRFE \cite{ref_rrfe} removes the batch normalization layer during incremental training to enhance model stability and employs prototype mixing to expand the feature space of old classes. In contrast to these methods, we propose modeling the high-dimensional feature space using multivariate Gaussian distributions to generate higher-quality representations of old classes, ensuring both greater diversity and accuracy of old class representations.

\section{Methodology}

\begin{figure*}[htb]
\centering
\includegraphics[width=\textwidth]{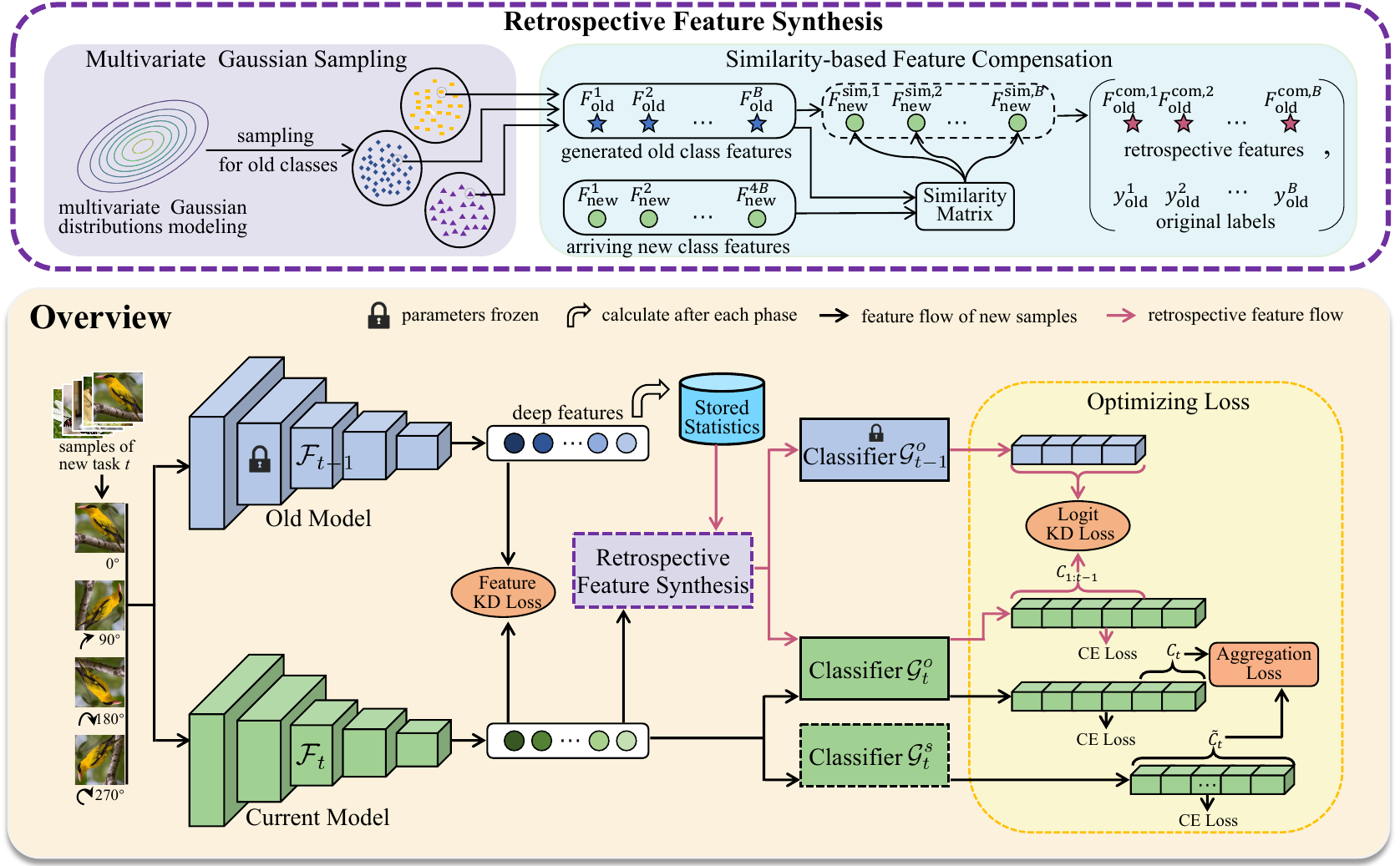}
\caption{\small Overall framework of our proposed method for NECIL. The new samples from the current task are augmented using rotation transformations. These augmented new class samples, along with synthesized retrospective features, are then utilized to optimize the incremental classification model to consolidate previously acquired knowledge while learning new classes. In the process of retrospective feature synthesis, we model the feature space of each old class using a multivariate Gaussian distribution and sample old class representations from high-likelihood regions. Additionally, we select the most similar new class features to complement the sampled old class features, thereby creating efficient retrospective representations of the old classes. ``CE loss" represents the cross-entropy classification loss. ``Stored Statistics" refers to the feature means and covariance matrices, which are calculated and stored after each task is learned. $\tilde{C}_{t}$ represents the classes within task $t$ after rotation-based label augmentation.}
\label{fig_our_pipeline}
\end{figure*}

\subsection{Problem Definition and Overall Framework} \label{sec:method_1}

NECIL focuses on developing the classification model that can incrementally learn new classes while preserving its ability to recognize all previously learned classes, without the need to store raw samples from past tasks. Let $C_{t}$ represent the set of classes learned during task $t$, and $D_{t} = \{X_{t}, Y_{t}\} = \{x_{t}^{i}, y_{t}^{i}\}_{i=1}^{N_{t}}$ denote the data for task $t$, where $x_{t}^{i}$ is an input sample, $y_{t}^{i} \in C_{t}$ is its corresponding label, and $N_{t}$ is the total number of samples in $D_{t}$. Classes learned across different tasks are distinct, i.e., $C_{1} \cap C_{2} \cap \cdots \cap C_{T} = \emptyset$. The classification model consists of two key components:  the feature extractor $\mathcal{F}$ with parameters $\theta$, and the unified classifier $\mathcal{G}$ with parameters $\varphi$. The goal for task $t$ is to minimize a predefined loss function $\mathcal{L}_{t}$ on the new dataset $D_{t}$:

{ \small
\begin{equation}
\begin{aligned}
&\underset{\theta_{t}, \varphi_{t},\delta}{\operatorname{argmin}} \
\mathcal{L}_{t}\left(\mathcal{G}\left(\mathcal{F}\left(X_{t}; \theta_{t}\right); \varphi_{t}\right), Y_{t}\right) + \sum \delta_{p} \\
\text{s.t. } \mathcal{L}_{t}&\left(X_{p}, Y_{p}\right) - \mathcal{L}_{p}\left(X_{p}, Y_{p}\right) \leqslant \delta_{p}, \delta_{p} \geqslant 0, \quad \forall p \in [1, t-1]
\end{aligned}
\end{equation}
}

\noindent where $\mathcal{L}_{t}(X_{p},Y_{p})=\mathcal{L}(\mathcal{G}(\mathcal{F}(X_{p};\theta_{t});\varphi_{t}),Y_{p})$ represents the loss of the current model at task $t$ on the old dataset $D_{p}$, while $\mathcal{L}_{p}(X_{p},Y_{p})=\mathcal{L}(\mathcal{G}(\mathcal{F}(X_{p};\theta_{p});\varphi_{p}),Y_{p})$ denotes the loss of the previous model trained during task $p$ on the same dataset. The last term, $\delta=\{\delta_{p}\}$, is a slack variable that allows for a small permissible increase in the loss on the old dataset.

Figure \ref{fig_our_pipeline} illustrates the overall framework of our approach. After each phase, we store statistical information for each learned class, specifically the mean and covariance matrix of deep features derived from the training data. This information is used in the subsequent incremental learning phase to synthesize retrospective features of previously learned classes. Specifically, the proposed multivariate Gaussian sampling strategy models the feature space using a multivariate Gaussian distribution, generating representations of old classes by sampling from high-likelihood regions. To enhance these generated features, we apply a similarity-based feature compensation mechanism that selects similar new class features to complement the generated old class representations to effectively synthetic retrospective features.

During incremental learning of new classes, the new data is augmented using rotation transformations before being passed into the feature extractor. The deep features extracted from these new instances are used to optimize both the classifier $\mathcal{G}^{s}_{t}$ with self-supervised label augmentation \cite{ref_ssl_la,ref_praka} and the standard non-augmented classifier $\mathcal{G}^{o}_{t}$. The augmented classifier $\mathcal{G}^{s}_{t}$ classifies only the classes learned in the current phase, treating each rotation transformation as a distinct class. In contrast, $\mathcal{G}^{o}_{t}$ is a unified classifier that recognizes the non-rotation-augmented classes, including both previously learned and newly learned classes. The synthesized retrospective features for the old classes are injected into $\mathcal{G}^{o}_{t}$ to reconstruct decision boundaries of these classes. To maintain model stability and prevent significant degradation on old classes when new data is introduced, we apply knowledge distillation \cite{ref_kd} at both the feature level and the classification logits. During the evaluation (inference) phase, the current feature extractor $\mathcal{F}_{t}$ and classifier $\mathcal{G}_{t}^{o}$ are used for testing.

\subsection{Retrospective Feature Synthesis}

\noindent \textbf{Multivariate Gaussian sampling.} Inspired by VOS \cite{ref_vos}, we assume that the representations of old classes follows a class-conditional multivariate Gaussian distribution:

\begin{equation}
    p(\mathcal{F}(x;\theta_{t}) | y = k) = \mathcal{N}(\boldsymbol{\mu}_{k}, \mathbf{\Sigma}_{k})
\end{equation}

\noindent where $\boldsymbol{\mu}_{k}$ is the mean of class $k \in \{1, 2, \ldots, T\}$, $\mathbf{\Sigma}_{k}$ is the tied covariance matrix, and $\mathcal{F}(x) \in \mathbb{R}^{m}$ represents the latent feature extracted from image $x$ by the feature extractor, with $m$ being the dimension of the deep feature space. To estimate the parameters of this Gaussian distribution, we compute the empirical mean ($\widehat{\boldsymbol{\mu}}_{k}$) and covariance ($\widehat{\boldsymbol{\Sigma}}_{k}$) from the high-dimensional features of all training samples:

\begin{equation}
    \begin{aligned}
        \widehat{\boldsymbol{\mu}}_{k} &= \frac{1}{N^{k}} \sum_{i: y_{i} = k} \mathcal{F}(x_{i}; \theta_{p}), \\
        \widehat{\boldsymbol{\Sigma}}_{k} &= \frac{1}{N^{k}} \sum_{i: y_{i} = k} (\mathcal{F}(x_{i}; \theta_{p}) - \widehat{\boldsymbol{\mu}}_{k})(\mathcal{F}(x_{i}; \theta_{p}) - \widehat{\boldsymbol{\mu}}_{k})^{\top}
    \end{aligned}
\end{equation}

\noindent where $N^{k}$ is the number of images in class $k$, and $p$ represents the incremental learning phase which class $k$ is learned in. After each phase of training is completed, the multivariate Gaussian parameters of the classes learned in this phase are evaluated.

We propose generating deep representations for old classes by sampling from the estimated multivariate Gaussian distribution. To ensure that the generated features have high confidence and diversity, we sample old class features $\mathcal{O}_{k}$ from the $\zeta$ likelihood region of the estimated class-conditional distribution:

{ \small
\begin{align}
    \mathcal{O}_{k} = \{\mathbf{o}_{k} \mid & \frac{1}{\sqrt{(2\pi)^{m} \widehat{\boldsymbol{\Sigma}}_{k}}} \exp\left( \frac{(\mathbf{o}_{k} - \widehat{\boldsymbol{\mu}}_{k})^{\top} (\mathbf{o}_{k} - \widehat{\boldsymbol{\mu}}_{k})}{-2 \widehat{\boldsymbol{\Sigma}}_{k}} \right)
     > \zeta \}
\end{align}
}

\noindent Here, $\mathcal{O}_{k} \sim \mathcal{N}(\boldsymbol{\mu}_{k}, \boldsymbol{\Sigma}_{k})$ represents the deep features sampled for class $k$, specifically from the sub-level set with likelihood greater than $\zeta$. These sampled features should ideally aid the classifier in reconstructing the decision boundaries of old classes. For consistency, we generate the same number of old class features as the batch size of new data. In practice, we implement this feature generation process by selecting the sample with the highest likelihood from a set of 1000 ($K=1000$) randomly sampled instances for each old class. We also conduct ablation experiments on different values of $K$ to verify its robustness (as shown in Figure \ref{fig:diff_k}). Each iteration generates different samples to ensure feature diversity, while the high-likelihood constraint ensures feature quality. Together, these guarantees enhance the classifier's ability to accurately reconstruct the decision boundaries of previous classes.

\noindent\textbf{Similarity-based feature compensation.} 
While the features generated in the previous step help retain information about previous classes, these features gradually become outdated as the model is updated incrementally \cite{ref_praka}. Specifically, maintaining the model’s plasticity requires updating the classifier when learning new classes. This update can lead to a discrepancy between the synthesized old class features and the current classifier, potentially resulting in a bias towards new classes. To address this, we compensate for the generated old class representations by incorporating new class features, which are better aligned with the updated classifier.

As depicted in Figure \ref{fig_our_pipeline}, we first rotate each new class sample $x_{t}^{i}$ by 0°, 90°, 180°, and 270° to generate augmented features $\tilde{x}_{t}^{i}$:
\begin{equation}
    \tilde{x}_{t}^{i} = \{ x_{t}^{i,j} \}_{j=0}^{3} = \{ \text{rotate}(x_{t}^{i}, j \times 90^{\circ}) \}_{j=0}^{3}
\end{equation}
A batch of these rotation-augmented images is then passed through the feature extractor to obtain the new class features $\tilde{F}_{\mathrm{new}}$.

Let $C_{o}$ denote the set of all old classes at current phase, and let $B$ represent the batch size. We then determine a batch of old classes, denoted as $y_{\mathrm{old}}$, as follows:

{ \small
\begin{equation}
    y_{\mathrm{old}} = \begin{cases}
        \text{rand}(C_{o}, B), & \text{if } B < \text{size}(C_{o}) \\
        C_{o}, & \text{if } B = \text{size}(C_{o}) \\
        C_{o} \cup \text{rand}(C_{o}, B - \text{size}(C_{o})), & \text{if } B > \text{size}(C_{o})
    \end{cases}
\end{equation}
}

\noindent where $\text{rand}(C_{o}, B)$ represents a random selection of $B$ classes from $C_{o}$. For each class in $y_{\mathrm{old}}$, a corresponding feature is generated using the MGS strategy, forming the generated feature set $F_{\mathrm{old}}$. The cosine similarity $S$ between the old class features $F_{\mathrm{old}}$ and the new class features $\tilde{F}_{\mathrm{new}}$ is calculated as follows:
\begin{equation}
    S \in \mathbb{R}^{B \times 4B} = \cos(F_\mathrm{old}, \tilde{F}_\mathrm{new})
\end{equation}

For each old class feature in $F_{\text{old}}$, the most similar new class features in $\tilde{F}_{\text{new}}$ is selected to form the similar feature set $F_{\text{new}}^{\text{sim}}$:

\begin{equation}
    F_{\text{new}}^{\text{sim}} = \left[\tilde{F}_{\text{new}, j^*_1}, \tilde{F}_{\text{new}, j^*_2}, \ldots, \tilde{F}_{\text{new}, j^*_B}\right]
\end{equation}

\noindent where $j^*_i = \arg\max_{j} \; S_{ij}, \quad \forall i \in \{1, 2, \ldots, B\}$. Element-wise averaging is then performed to obtain the compensated old class features $F_{\text{old}}^{\text{com}}$:

\begin{equation}
    F_{\text{old}}^{\text{com}} = \frac{F_{\text{old}} + F_{\text{new}}^{\text{sim}}}{2}
\end{equation}

\noindent $F_{\text{old}}^{\text{com}}$ and the original labels $y_{\text{old}}$ are paired as a batch of synthesized information injected into the classification model. Since the compensated old class features are similar to the new class features while retaining the characteristics of the old classes, they better enhance the model’s discriminative ability during the joint learning process.


\subsection{Optimizing Loss}
\noindent\textbf{New classes learning.} For the rotation-augmented image $\tilde{x}_{t}^{i}$, the augmented labels are defined as:
\begin{equation}
    \tilde{y}_{t}^{i} = \left\{ y_{t}^{i} \times 4 + j \right\}_{j=0}^{3}
\end{equation}
We first optimize the two classifiers separately using cross-entropy classification loss. For the classifier $\mathcal{G}^{s}_{t}$, the cross-entropy classification loss $\mathcal{L}_{\text{new}}^{\text{aug\_cls}}$ is computed using the augmented data $\tilde{x}_{t}$ and labels $\tilde{y}_{t}$. For the standard classifier $\mathcal{G}^{o}_{t}$, the unaugmented data (0°) $x_{t}$ and original labels $y_{t}$ are used to compute the cross-entropy classification loss $\mathcal{L}_{\text{new}}^{\text{cls}}$.

To further enhance model generalization on arriving new classes, we employ Kullback-Leibler divergence (KL) loss to compute the aggregation loss between the outputs of these two classifiers, following PRAKA \cite{ref_praka}:
\begin{equation}
    \mathcal{L}^{\text{ka}}_{\text{new}} = \text{KL}(\mathcal{P}_{\text{agg}}(\tilde{y}_{t}^{i} \mid \tilde{F}_{\text{new}}) || \mathcal{P}(y_{t}^{i} \mid F_{\text{new}}))
\end{equation}

\noindent where $F_{\text{new}}$ represents the hidden features obtained by the feature extractor on the non-augmented image $x_{t}$. $\mathcal{P}_{\text{agg}}(\tilde{y}^{i}_{t} \mid \tilde{F}_{\text{new}})$ represents the aggregated outputs of classifier $\mathcal{G}^{s}_{t}$, and $\mathcal{P}(y^{i}_{t} \mid F_{\text{new}})$ represents the output of the classifier $\mathcal{G}^{o}_{t}$ on the current learning classes. The definitions of these two components are as follows:
\begin{equation}
    \mathcal{P}_{\text{agg}}(\tilde{y}^{i}_{t} \mid \tilde{F}_{\text{new}}) = \frac{1}{4} \sum_{j=0}^{3} {\varphi}_{t,s}^{j} \tilde{F}_{\text{new}}^{i,j}
\end{equation}
\begin{equation}
    \mathcal{P}(y^{i}_{t} \mid F_{\text{new}}) = \mathbb{I}_{C_{t}}({\varphi}_{t,o} F_{\text{new}}^{i})
\end{equation}

\noindent where $\mathbb{I}_{C_{t}}$ is an indicator function that filters the output to only the current classes in $C_{t}$. Thus, the loss for learning new classes is summarized as follows:
\begin{equation}
    \begin{aligned}
    \mathcal{L}_{\text{new}} =& \mathcal{L}_{\text{new}}^{\text{cls}} + \mathcal{L}_{\text{new}}^{\text{aug\_cls}} + \mathcal{L}^{\text{ka}}_{\text{new}}
\end{aligned}
\end{equation}

\noindent\textbf{Retrospective knowledge retention.}
Similarly, the synthesized representations of old classes $F_{\text{old}}^{\text{com}}$ and their corresponding labels $y_{\text{old}}$ are used to optimize the classifier $\mathcal{G}^{o}_{t}$, reconstructing the decision boundaries for old classes through cross-entropy classification loss:

\begin{equation}
    \mathcal{L}_{\text{old}}^{\text{cls}} = \mathcal{L}_{\text{ce}}(\mathcal{G}^{o}(F_{\text{old}}^{\text{com},i}; \varphi_{t,o}), y_{\text{old}}^{i})
\end{equation}
Additionally, the synthesized old class representations and arriving new class features are used for knowledge distillation at both the logit output level and the deep feature level to prevent significant degradation in model stability. The feature-level knowledge distillation, as used in PASS \cite{ref_pass}, RRFE \cite{ref_rrfe}, and PRAKA \cite{ref_praka}, is defined as:
\begin{equation}
    \mathcal{L}_{\text{old}}^{\text{feat\_kd}} = \| \mathcal{F}(\tilde{x}_{t}^{i}; \theta_{t}) - \mathcal{F}(\tilde{x}_{t}^{i}; \theta_{t-1}) \|_{2}
    \label{eq:feat_dist}
\end{equation}
The logit-level knowledge distillation is defined as:
{\small
\begin{equation}
    \mathcal{L}_{\text{old}}^{\text{logit\_kd}} = \text{KL}(\mathcal{G}^{o}(F_{\text{old}}^{\text{com},i}; \varphi_{t,o}),\mathcal{G}^{o}(F_{\text{old}}^{\text{com},i}; \varphi_{t-1,o}))
    \label{eq:logit_dist}
\end{equation}
}
Thus, the loss for retaining retrospective knowledge is summarized as:
\begin{equation}
    \mathcal{L}_{\text{old}} = \mathcal{L}_{\text{old}}^{\text{cls}} + \mathcal{L}_{\text{old}}^{\text{feat\_kd}} + \mathcal{L}_{\text{old}}^{\text{logit\_kd}}
\end{equation}

\noindent\textbf{Overall optimization loss.}
Combining the two parts, the overall optimization objective of our proposed method on the current task $t$ is defined as:
\begin{equation}
    \mathcal{L}_{t} = \mathcal{L}_{\text{new}} + \alpha \mathcal{L}_{\text{old}}
\end{equation}
where $\alpha$ is a hyperparameter to balance the weights of these two components,  and we set it to 15 in our experiments following PRAKA \cite{ref_praka}.

\section{Experiments and Analysis}

\noindent\textbf{Datasets.} 
We evaluate the performance of our method through expensive experiments on three datasets: CIFAR-100 \cite{ref_cifar100}, TinyImageNet \cite{ref_tiny_imagenet}, and ImageNet-Subset \cite{ref_imagenet_subset}. CIFAR-100 contains 100 classes, a total of 60,000 32×32 pixel images, with 500 images for training and 100 for testing in each class. TinyImageNet includes 200 classes, each containing 500 training images, 50 validation images, and 50 test images, offering a more challenging scene for performance comparison. The ImageNet-Subset is a 100-class subset of ImageNet-1k, which has larger image sizes and scales than CIFAR-100 and TinyImageNet. We employ an incremental learning setup denoted as B + C × T. Initially, the model is trained on B classes. The remaining classes are then evenly distributed across T subsequent phases (tasks), with C classes per phase. B is set to half the total number of classes, with minor adjustments made to ensure the remaining classes are evenly divided across the subsequent phases. We determine three different incremental learning settings on each dataset by varying the T value (5, 10, 20). For simplicity, these settings are referred to as 5 phases, 10 phases, and 20 phases. To fairly compare performance, the class splits for each incremental setting on all datasets strictly follow those used in \cite{ref_pass} and \cite{ref_praka}.

\noindent\textbf{Implementation details.}
Following \cite{ref_pass,ref_ssre,ref_praka,ref_rrfe}, we use ResNet-18 as the backbone network (feature extractor). Each new task is trained for 100 epochs, using the Adam optimizer with parameters $\beta_1 = 0.9$, $\beta_2 = 0.999$, and a weight decay of 2e-4. The batch size is set to 128, and the initial learning rate is 0.001, which is reduced by a factor of 10 at the 45 and 90 epochs. Results marked with an asterisk (*) are cited from \cite{ref_praka} or \cite{ref_rrfe}. To minimize the impact of training randomness on model accuracy, we fix the random seed to 0 during training, consistent with PRAKA \cite{ref_praka}. Additionally, we conduct experiments using different class orders and random seeds, as shown in Figure \ref{fig:rand_order_seed}, to verify the robustness of our method.

\noindent\textbf{Evaluation metrics.}
Consistent with \cite{ref_pass,ref_praka,ref_rrfe}, we use average incremental accuracy, final accuracy, and average forgetting as evaluation metrics. Average incremental accuracy measures the average classification accuracy of the model across all incremental stages (including the initial task), reflecting the method's overall incremental performance. Final accuracy represents the top-1 classification accuracy achieved after completing all incremental learning phases. Average forgetting, as defined in \cite{ref_forget_metric}, quantifies the average difference between the peak accuracy and final accuracy for each task after incremental learning. A smaller difference indicates better performance.

\begin{table*}[htbp]
\centering
\caption{\small Quantitative comparison of average incremental accuracy (\%) between our method and other methods on CIFAR-100, TinyImageNet, and ImageNet-Subset. E represents the number of retained exemplars. Best results are highlighted in bold.}
\setlength\tabcolsep{1.2mm}
\resizebox{\linewidth}{!}{
\begin{tabular}{l|l|ccc|ccc|ccc}
\toprule
\multicolumn{2}{c|}{\multirow{2}{*}{\textbf{Methods}}} & \multicolumn{3}{c|}{\textbf{CIFAR-100}} & \multicolumn{3}{c|}{\textbf{TinyImageNet}} & \multicolumn{3}{c}{\textbf{ImageNet-Subset}}   \\
\multicolumn{2}{l|}{}                         & \textbf{5 phases}      & \textbf{10  phases}     & \textbf{20  phases}    & \textbf{5  phases}       & \textbf{10  phases}      & \textbf{20  phases}     & \textbf{5  phases}       & \textbf{10  phases}     & \textbf{20  phases}   \\ \midrule
\multirow{5}{*}{\rotatebox{90}{\textit{E = 20}}} & iCaRL-CNN$^{\ast}$      & 51.07    & 48.66    & 44.43   & 34.64     & 31.15     & 27.90& 53.62     & 50.53    & —      \\
                                                 & iCaRL-NCM$^{\ast}$  \cite{ref_icarl}    & 58.56    & 54.19    & 50.51   & 45.86     & 43.29     & 38.04    & 65.04     & 60.79    & —      \\
                                                 & EEIL$^{\ast}$   \cite{ref_efil}        & 60.37    & 56.05    & 52.34   & 47.12     & 45.01     & 40.50    & —         & 63.34    & —      \\
                                                 & UCIR$^{\ast}$   \cite{ref_ucir}        & 63.78    & 62.39    & 59.07   & 49.15     & 48.52     & 42.83    & 68.43     & 66.16    & —      \\
                                                 & WA$^{\ast}$     \cite{ref_wa}        & 64.32    & 59.51    & 54.37   & —         & —         & —        & —         & 63.60    & —      \\ \midrule
\multirow{11}{*}{\rotatebox{90}{\textit{E = 0}}} & EWC$^{\ast}$    \cite{ref_ewc}        & 24.48   & 21.20     & 15.89   & 18.80     & 15.77     & 12.39    & —         & 20.40& —      \\
                                                 & LwF-MC$^{\ast}$ \cite{ref_icarl}        & 45.93    & 27.43    & 20.07   & 29.12     & 23.10     & 17.43    & —         & 31.18    & —      \\
                                                 & MUC$^{\ast}$   \cite{ref_muc}         & 49.42    & 30.19    & 21.27   & 32.58     & 26.61     & 21.95& —         & 35.07    & —      \\
                                                 & SDC$^{\ast}$   \cite{ref_sdc}         & 56.77    & 57.00    & 58.90   &  —        & —         & —        & —         & 61.12    & —      \\
                                                 & IL2A$^{\ast}$  \cite{ref_il2a}         & 66.03    & 60.31    & 57.92   & 47.34     & 44.70     & 40.04    & —         & —        & —      \\
                                                 & PASS$^{\ast}$   \cite{ref_pass}        & 63.47    & 61.84    & 58.09   & 49.55     & 47.29     & 42.07    & 64.41     & 61.80    & 51.32  \\
                                                 & SSRE$^{\ast}$   \cite{ref_ssre}        & 65.88    & 65.04    & 61.70   & 50.39     & 48.93     & 48.17    & —         & 67.69    & —      \\
                                                 & FeTrIL$^{\ast}$ \cite{ref_fetril}        & 64.73    & 63.41    & 57.44   & 52.91     & 51.71     & 49.72    & 69.64     & 68.93    & 62.52  \\
                                                 & RRFE$^{\ast}$  \cite{ref_rrfe}         & 67.56    & 66.76    & 64.32   & 51.50     & 50.99     & 50.56    & 69.56     & 69.05    & 63.73  \\
                                                 & PRAKA$^{\ast}$ \cite{ref_praka}         & 70.02    & 68.86    & 65.86   & \textbf{53.32}     & 52.61     & 49.83    & —         & 68.98    & —      \\
                                                 & RFS (Ours)       & \textbf{71.82}    & \textbf{71.57}    & \textbf{69.39}   & 53.25     & \textbf{53.68}     & \textbf{52.68}    & \textbf{74.15}     & \textbf{73.87}    & \textbf{72.43}  \\ \bottomrule
\end{tabular}}
\label{tab:average_incre_acc}
\end{table*}

\begin{table}[!htbp]
\centering
\caption{ \small Comparison of final accuracy (\%) between our method and state-of-the-art methods on CIFAR-100 and TinyImageNet.}
\resizebox{0.48\textwidth}{!}{
\begin{tabular}{@{}l|ccc|ccc@{}}
\toprule
\multicolumn{1}{c|}{\multirow{2}{*}{\textbf{Methods}}} & \multicolumn{3}{c|}{\textbf{CIFAR-100}}           & \multicolumn{3}{c}{\textbf{TinyImageNet}}        \\
\multicolumn{1}{c|}{}                                  & \textbf{5}   & \textbf{10} & \textbf{20} & \textbf{5}  & \textbf{10} & \textbf{20} \\ \midrule
PASS                                                  & 56.49          & 50.32          & 47.02          & 42.03          & 40.35          & 34.30\\
SSRE                                                  & 56.91          & 56.45          & 51.54          & 44.56          & 44.02          & 43.77          \\
PRAKA                                                 & 61.79          & 60.44          & 56.31          & 46.19          & 45.82          & 41.09          \\
RFS (Ours)                                                  & \textbf{64.22} & \textbf{64.21} & \textbf{60.41} & \textbf{47.13} & \textbf{47.86} & \textbf{45.23} \\ \bottomrule
\end{tabular}}
\label{tab:final_acc}
\end{table}

\begin{figure*}[htbp]
\centering
\includegraphics[width=\textwidth]{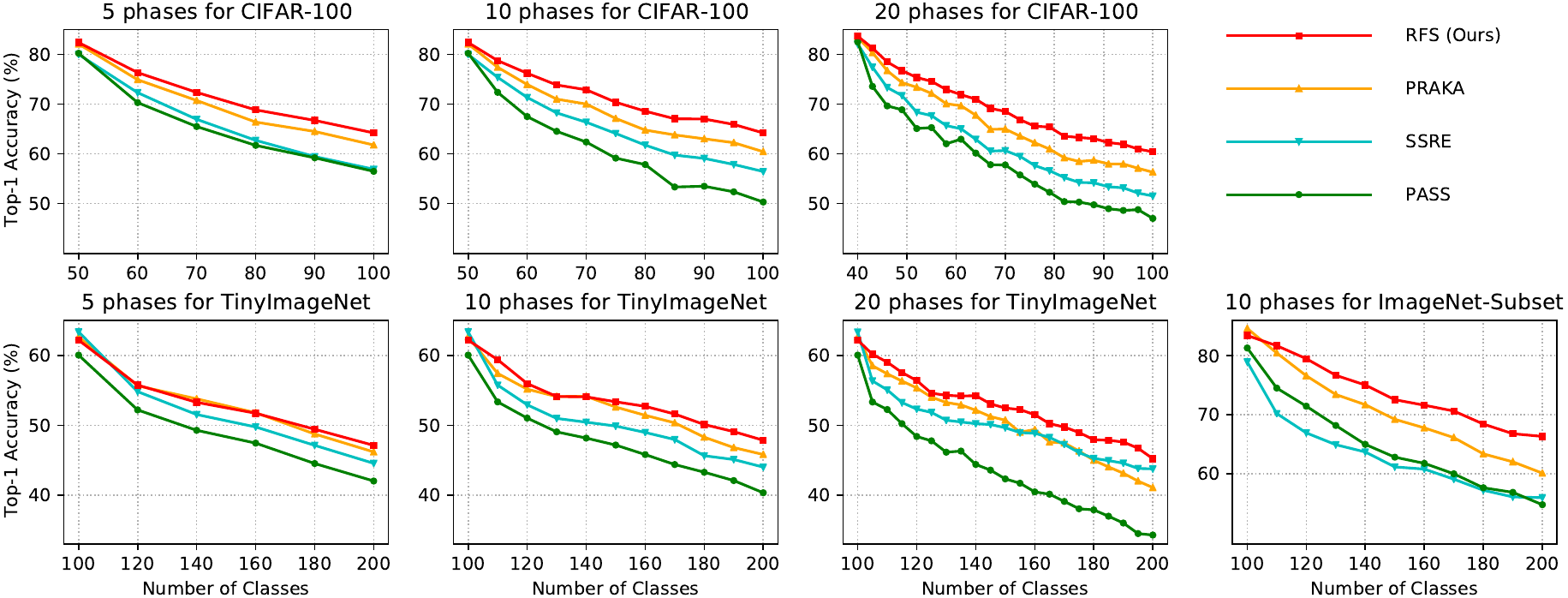}
\caption{\small Average classification accuracy of our method and state-of-the-art methods at each phase during incremental learning on CIFAR-100 and TinyImageNet.}
\label{fig:acc_curve_compara}
\end{figure*}

\begin{table}[htbp]
\centering
\caption{\small Comparison of average forgetting between our method and other methods on CIFAR-100 and TinyImageNet.}
\resizebox{0.48\textwidth}{!}{
\begin{tabular}{@{}l|ccc|ccc@{}}
\toprule
\multirow{2}{*}{\textbf{Methods}} & \multicolumn{3}{c|}{\textbf{CIFAR-100}}          & \multicolumn{3}{c}{\textbf{TinyImageNet}}       \\
                                  & \textbf{5}    & \textbf{10}    & \textbf{20}    & \textbf{5}    & \textbf{10}    & \textbf{20}    \\ \midrule
iCARL-CNN$^{\ast}$                         & 42.13         & 45.69          & 43.54          & 36.89         & 36.70          & 45.12 \\
iCaRL-NCM$^{\ast}$                         & 24.90         & 28.32          & 35.53          & 27.15         & 28.89          & 37.40 \\
EFIL$^{\ast}$                              & 23.36         & 26.65          & 32.40          & 25.56         & 25.51          & 35.04 \\
UCIR$^{\ast}$                              & 21.00         & 25.12          & 28.65          & 20.61         & 22.25          & 33.74 \\ \midrule
LwF\_MC$^{\ast}$                           & 44.23         & 50.47          & 55.46          & 54.26         & 54.37          & 63.54 \\
MUC$^{\ast}$                               & 40.28         & 47.56          & 52.65          & 51.46         & 50.21          & 58.00 \\
PASS$^{\ast}$                              & 25.20         & 30.25          & 30.61          & 18.04         & 23.11          & 30.55 \\
SSRE$^{\ast}$                              & 18.37         & 19.48          & 19.00          & 9.17          & 14.06          & 14.20 \\
RRFE$^{\ast}$                              & 16.02         & 15.57          & 17.43          & 13.49         & 14.26          & 13.18 \\
PRAKA$^{\ast}$                             & 12.59         & 14.65          & 17.39          & 11.84         & 13.95          & 18.51 \\
RFS (Ours)                              & \textbf{5.76}    & \textbf{5.04}  & \textbf{7.14}  & \textbf{4.06} & \textbf{3.35}  & \textbf{6.58} \\ \bottomrule      
\end{tabular}}
\label{tab:forgetting}
\end{table}

\begin{table}[htbp]
\centering
\caption{Ablation study (in average incremental accuracy, \%) of our method on CIFAR-100 and TinyImageNet.}
\resizebox{0.48\textwidth}{!}{
\begin{tabular}{@{}cc|ccc|ccc@{}}
\toprule
\multirow{2}{*}{\textbf{MGS}} & \multirow{2}{*}{\textbf{SFC}} & \multicolumn{3}{c|}{\textbf{CIFAR-100}}                                                                        & \multicolumn{3}{c}{\textbf{TinyImageNet}}                                                                    \\
                              &                               & \textbf{5}      & \textbf{10}       & \textbf{20}       & \textbf{5}        & \textbf{10}       & \textbf{20}       \\ \midrule
                              &                               & 60.28          & 55.00          & 47.62          & 44.68          & 40.10          & 32.51          \\
\checkmark                    &                               & 69.47          & 67.36          & 62.87          & 51.17          & 49.25          & 43.90           \\
                              & \checkmark                    & 70.26          & 68.18          & 63.71          & 52.83          & 52.02          & 48.18          \\
\checkmark                    & \checkmark                    & \textbf{71.82} & \textbf{71.57} & \textbf{69.39} & \textbf{53.25} & \textbf{53.68} & \textbf{52.68} \\ \bottomrule
\end{tabular}}
\label{tab:main_abla}
\end{table}

\begin{figure}[htbp]
\centering
\includegraphics[width=0.48\textwidth]{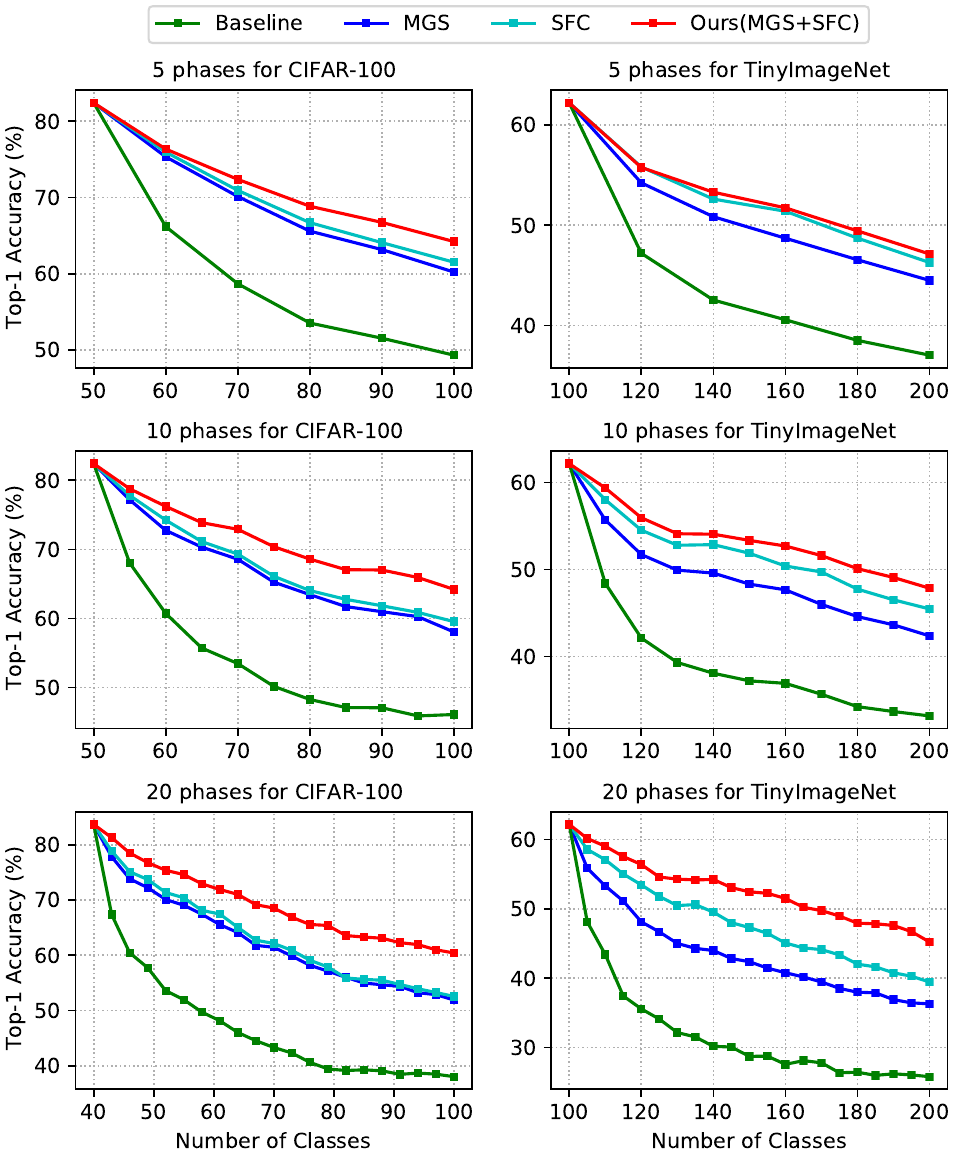}
\caption{\small Average classification accuracy for the component effectiveness analysis of our method on CIFAR-100 and TinyImageNet.}
\label{fig:abla_acc_curve}
\end{figure}

\subsection{Overall Performance} \label{sec_overall}

Table \ref{tab:average_incre_acc} presents a quantitative comparison of average incremental accuracy between various methods. Compared to the listed methods that store previous exemplars (\textit{E=20}), our method demonstrates significantly improved average incremental accuracy across different incremental settings on all three datasets. Moreover, our method also outperforms non-exemplar methods (\textit{E=0}), achieving the best overall performance. Notably, compared to the state-of-the-art PRAKA method, our method achieves average accuracy improvements of 1.8\%, 2.71\%, and 3.53\% in incremental settings of 5, 10, and 20 phases on CIFAR-100, respectively. On TinyImageNet, our method achieves accuracy gains of 1.07\% and 2.85\% in the 10 and 20 phases settings. On ImageNet-Subset, our method shows an average improvement of 4.89\% in the 10 phases setting. Additionally, as the number of incremental phases increases, our method demonstrates greater robustness than PRAKA, effectively mitigating the rapid decline in average accuracy. These results highlight the significant superiority of our method in overall incremental learning performance.

Table \ref{tab:final_acc} further compares the final accuracy of our method with several state-of-the-art approaches after completing different incremental learning settings. Our method significantly outperforms PASS and SSRE, and shows accuracy improvements of 2.43\%, 3.77\%, and 4.1\% over PRAKA in the three incremental settings on CIFAR-100. In the three settings on TinyImageNet, our method also achieves accuracy gains of 0.94\%, 2.04\%, and 4.14\%, respectively. This demonstrates that our method has superior visual recognition performance after all incremental phases. Figure \ref{fig:acc_curve_compara} shows the classification accuracy curves of our method compared to these state-of-the-art methods at different incremental phases. Our method consistently maintains higher accuracy in nearly all phases, with a slower decline in accuracy as the number of incremental phases increases. To further differentiate between the methods, we compare their average forgetting at the last phase in Table \ref{tab:forgetting}. Our method achieves lower average forgetting rates in the 5, 10, and 20 phases settings on both CIFAR-100 and TinyImageNet.

\subsection{Ablation Studies}
In this section, we perform ablation studies to analyze the impact of the key components of our proposed method: Multivariate Gaussian Sampling (MGS) and Similarity-based Feature Compensation (SFC).

As illustrated in the first two rows of Table \ref{tab:main_abla}, incorporating MGS significantly enhances the average incremental accuracy across various incremental settings on CIFAR-100 and TinyImageNet, compared to the method without MGS. Similarly, the last two rows demonstrate that the combined use of MGS and SFC leads to further performance gains over using SFC alone. Figure \ref{fig:abla_acc_curve} illustrates that MGS substantially improves classification accuracy at each phase, resulting in a slower decline in accuracy compared to the baseline. These findings emphasize the pivotal role of MGS in non-exemplar class-incremental learning.

Furthermore, as indicated in the first and third rows of Table \ref{tab:main_abla}, incorporating SFC also leads to significant improvements in average incremental accuracy across both datasets. The second and fourth rows show that the combined use of SFC and MGS yields substantial performance improvements compared to using MGS alone. In Figure \ref{fig:abla_acc_curve}, SFC enhances classification accuracy at all phases of incremental learning compared to both the baseline and MGS alone, again showing a slower decline in accuracy. These results highlight the substantial impact of SFC and demonstrate that MGS and SFC complement each other, with their combined use delivering the most significant improvements in NECIL.

\subsection{Analysis}

\begin{figure*}[htbp]
\centering
\captionsetup[subfigure]{font=footnotesize, labelfont={footnotesize}} 
\hfill
\subfloat[\scriptsize Visualizations for MGS]{\includegraphics[width=0.48\textwidth]{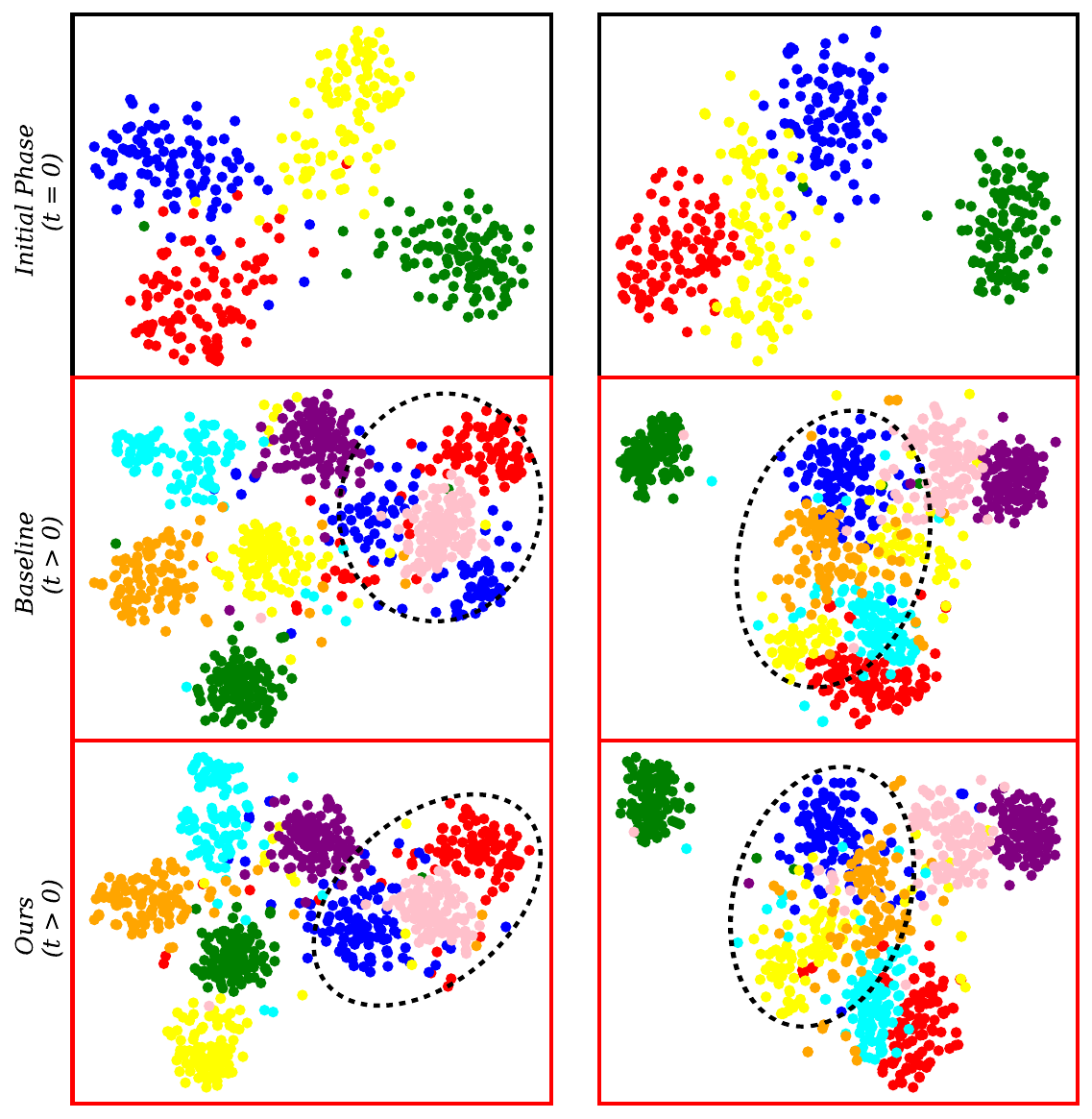}
\label{fig:visualization_dfg}}
\hspace{0px}
\subfloat[\scriptsize Visualizations for SFC]{\includegraphics[width=0.48\textwidth]{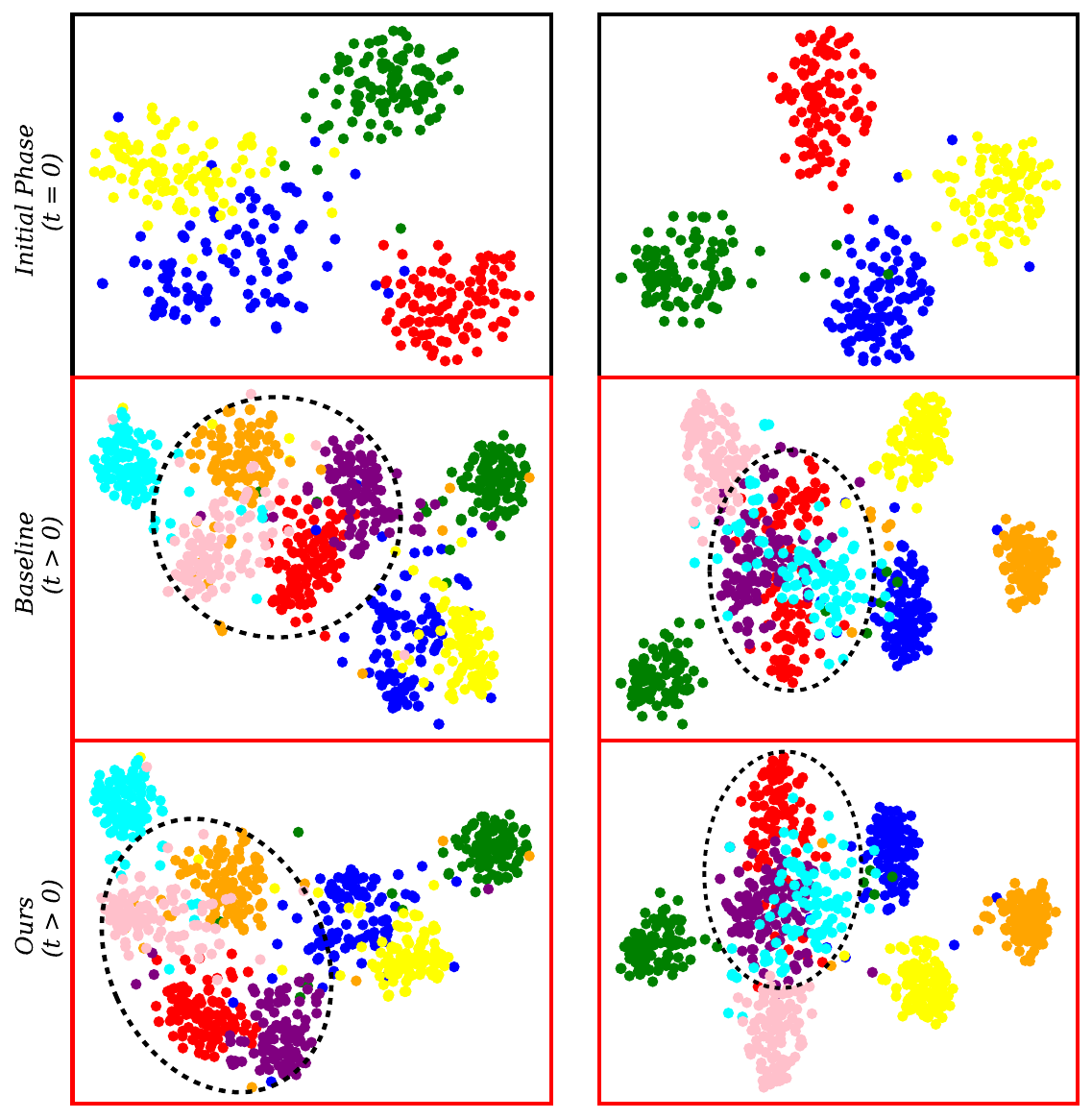}
\label{fig:visualization_sfc}}
\hfill
\caption{\small Visualization results. (a) MGS enhances the clustering of old classes in deep feature space. (b) SFC increases the discrimination between old and new classification features.}
\label{fig:visualization}
\end{figure*}

\noindent \textbf{Visualization results.} To intuitively demonstrate the effectiveness of the different modules in our method, we use t-SNE \cite{ref_tsne} to visualize the 2D embeddings of deep feature vectors, as shown in Figure \ref{fig:visualization}. The two columns in Figure \ref{fig:visualization_dfg} show that MGS effectively reduces internal distances of old classes compared to the baseline method. This enhancement improves the clustering of old class distributions and mitigates the model's bias towards new classes. In Figure \ref{fig:visualization_sfc}, the first column illustrates that SFC increases cluster separation between old and new classes. Additionally, the second column shows that SFC reduces feature confusion between old and new classes. Together, these results demonstrate the effectiveness of SFC in improving the model's ability to distinguish between new and old classes.

\begin{table*}[htbp]
\centering
\caption{Quantitative comparison of average incremental accuracy (Avg,\%) and final accuracy (Final, \%) between our method and other feature generation methods.}
\setlength\tabcolsep{1.2mm}
\resizebox{0.85\textwidth}{!}{
\begin{tabular}{l|cc|cc|cc|cc|cc|cc}
\toprule
\multirow{3}{*}{\textbf{Methods}} & \multicolumn{6}{c|}{\textbf{CIFAR-100}}                                                                                  & \multicolumn{6}{c}{\textbf{TinyImageNet}}                                                                            \\ \cline{2-13}
                                  & \multicolumn{2}{c|}{\textbf{5 phases}} & \multicolumn{2}{c|}{\textbf{10 phases}} & \multicolumn{2}{c|}{\textbf{20 phases}} & \multicolumn{2}{c|}{\textbf{5 phases}} & \multicolumn{2}{c|}{\textbf{10 phases}} & \multicolumn{2}{c}{\textbf{20 phases}} \\ 
                                  & Avg↑               & Final↑             & Avg↑              & Final↑             & Avg↑              & Final↑             & Avg↑              & Final↑            & Avg↑              & Final↑             & Avg↑              & Final↑             \\ \hline
Prototype                & 70.26          & 61.51          & 68.18          & 59.54          & 63.71          & 52.55          & 52.83          & 46.28          & 52.02          & 45.48          & 48.18          & 39.46          \\
Prototype Mixing \cite{ref_rrfe}         & 68.87          & 59.88          & 67.89          & 58.79          & 63.69          & 51.83          & 51.65          & 44.65          & 50.95          & 43.91          & 47.78          & 37.42          \\
Gaussian Noise Aug \cite{ref_pass}      & 70.84          & 62.23          & 69.18          & 60.77          & 65.29          & 54.08          & \textbf{53.44} & 46.52          & 52.82          & 45.82          & 49.87          & 41.37          \\
MGS (Ours)& \textbf{71.82} & \textbf{64.22} & \textbf{71.57} & \textbf{64.21} & \textbf{69.39} & \textbf{60.41} & 53.25          & \textbf{47.13} & \textbf{53.68} & \textbf{47.86} & \textbf{52.68} & \textbf{45.23} \\ \bottomrule
\end{tabular}}
\label{tab:different_fg}
\end{table*}

\begin{table*}[htbp]
\centering
\caption{Quantitative comparison of average incremental accuracy (Avg,\%) and final accuracy (Final, \%) between our method and other feature compensation methods.}
\setlength\tabcolsep{1.2mm}
\resizebox{0.85\textwidth}{!}{
\begin{tabular}{l|cc|cc|cc|cc|cc|cc}
\toprule
\multirow{3}{*}{\textbf{Methods}} & \multicolumn{6}{c|}{\textbf{CIFAR-100}}                                                                                  & \multicolumn{6}{c}{\textbf{TinyImageNet}}                                                                            \\ \cline{2-13}
                                  & \multicolumn{2}{c|}{\textbf{5 phases}} & \multicolumn{2}{c|}{\textbf{10 phases}} & \multicolumn{2}{c|}{\textbf{20 phases}} & \multicolumn{2}{c|}{\textbf{5 phases}} & \multicolumn{2}{c|}{\textbf{10 phases}} & \multicolumn{2}{c}{\textbf{20 phases}} \\ 
                                  & Avg↑               & Final↑             & Avg↑              & Final↑             & Avg↑              & Final↑             & Avg↑              & Final↑            & Avg↑              & Final↑             & Avg↑              & Final↑             \\ \hline
None                              & 69.47              & 60.22             & 67.36             & 58.04             & 62.87             & 51.91             & 51.17             & 44.50             & 49.25             & 42.38             & 43.90              & 36.31             \\
Rand Interp \cite{ref_praka}                       & \textbf{72.19}     & \textbf{64.31}    & 71.37             & 63.54             & 68.12             & 57.64             & \textbf{53.58}    & \textbf{47.24}   & 53.06             & 46.53             & 50.29             & 41.99             \\
Rand Avg (Ours)                          & 71.78              & 64.04             & 71.35             & 63.98             & 69.14             & 59.37             & 53.18             & 46.57            & 53.26             & 47.15             & 52.40              & 45.09             \\
Least Sim Avg (Ours)                     & 69.49              & 59.45             & 68.84             & 60.34             & 67.11             & 57.16             & 50.08             & 42.80             & 49.58             & 42.79             & 48.46             & 40.43             \\
SFC (Ours)& 71.82              & 64.22             & \textbf{71.57}    & \textbf{64.21}    & \textbf{69.39}    & \textbf{60.41}    & 53.25             & 47.13            & \textbf{53.68}    & \textbf{47.86}    & \textbf{52.68}    & \textbf{45.23}  \\ \bottomrule 
\end{tabular}}
\label{tab:diff_fc}
\end{table*}

\begin{table*}[htbp]
\centering
\caption{Ablation study of our method with and without logit distillation on average incremental accuracy (Avg, \%) and final accuracy (Final, \%).}
\setlength\tabcolsep{1.2mm}
\resizebox{0.85\textwidth}{!}{
\begin{tabular}{c|cc|cc|cc|cc|cc|cc}
\toprule
\multirow{3}{*}{\textbf{Methods}} & \multicolumn{6}{c|}{\textbf{CIFAR-100}}                                                                                  & \multicolumn{6}{c}{\textbf{TinyImageNet}}                                                                            \\ \cline{2-13}
                                  & \multicolumn{2}{c|}{\textbf{5 phases}} & \multicolumn{2}{c|}{\textbf{10 phases}} & \multicolumn{2}{c|}{\textbf{20 phases}} & \multicolumn{2}{c|}{\textbf{5 phases}} & \multicolumn{2}{c|}{\textbf{10 phases}} & \multicolumn{2}{c}{\textbf{20 phases}} \\ 
                                  & Avg↑               & Final↑             & Avg↑              & Final↑             & Avg↑              & Final↑             & Avg↑              & Final↑            & Avg↑              & Final↑             & Avg↑              & Final↑             \\ \hline
Ours(w/o logit   distill)         & 71.68              & 64.05             & 71.49             & 63.99             & 69.19             & 60.11             & 53.14             & 46.97            & 53.58             & 47.50              & 52.41             & 44.87             \\
Ours(w/ logit distill)            & \textbf{71.82}     & \textbf{64.22}    & \textbf{71.57}    & \textbf{64.21}    & \textbf{69.39}    & \textbf{60.41}    & \textbf{53.25}    & \textbf{47.13}   & \textbf{53.68}    & \textbf{47.86}    & \textbf{52.68}    & \textbf{45.23}   \\ \bottomrule
\end{tabular}}
\label{tab:logit_dist}
\end{table*}

\noindent \textbf{Comparison with other feature generation methods.} 
To more comprehensively demonstrate the superiority of MGS, we compare its performance with different feature generation methods in Table \ref{tab:different_fg}. In this table, ``Prototype" refers to using prototypes as representations of old classes in high-dimensional feature space. ``Prototype Mixing" denotes the method employed in RREF, where prototypes of old classes are randomly and linearly combined to create hybrid prototypes with composite labels. ``Gaussian Noise Aug" refers to the method used in PASS, which involves augmented prototypes with random Gaussian noise.

The results indicate that our proposed MGS achieves more substantial performance improvements than both Prototype and Prototype Mixing. Compared to Gaussian Noise Aug, MGS shows average precision gains of 0.98\%, 2.39\%, and 4.1\% in the three incremental settings on CIFAR-100. For the 10-phase and 20-phase settings on TinyImageNet, MGS achieves average accuracy improvements of 0.86\% and 2.81\%, respectively. Additionally, MGS demonstrates superior performance in final accuracy after completing all incremental tasks, with more significant accuracy gains in longer-phase incremental learning settings. These results highlight the superiority of MGS, particularly in the incremental setting with longer phases.

\noindent \textbf{Comparison with other feature compensation methods.} To comprehensively demonstrate the superiority of SFC, we compare its performance with various feature compensation methods in Table \ref{tab:diff_fc}. ``None" refers to not compensating generated old class features. ``Rand Interp" refers to the random bidirectional interpolation between extracted new class features and generated old class features, as used in PRAKA to enrich old class representations. ``Rand Avg" randomly selects new class features to average with generated old class features, while ``Least Sim Avg" chooses new class features with the least similarity to average with generated old class features.
Compared to the method without feature compensation, the various compensation methods achieve significant improvements in overall performance, verifying the critical role of feature compensation in mitigating the deviation between old class features and the updated classifier. Additionally, SFC outperforms Rand Avg and Least Sim Avg across all incremental settings and metrics on both CIFAR-100 and TinyImageNet. This superior performance is due to more similar new class features being used for compensating the generated old class features, effectively reducing the discrepancy between old class features and the updated feature extractor.

Furthermore, while Rand Interp shows slightly better performance than Rand Avg in the 5 phases setting, Rand Avg demonstrates significantly improved accuracy in the 20 phases setting on the CIFAR-100 and TinyImageNet as the phase number of incremental setting increases. This suggests that the averaging-based feature compensation method performs better than the random interpolation-based method in longer incremental sequences. Random feature interpolation can lead to unstable adjustments in old class features, either reducing the difference with new class distributions or increasing the deviation from the true class distribution excessively. In contrast, the feature averaging maintains a relatively stable distinction between compensated features and new class features, thereby enhancing the model's discriminative ability.
Our method, SFC, combines similarity-based feature selection with feature averaging. It not only maintains a stable distinction between compensated and new class features through the consistent approach of feature averaging but also better alleviates the deviation between old class representations and the feature extractor by selecting the most similar new class features. As demonstrated by the experimental results, SFC outperforms other methods and achieves the best overall performance.

\begin{figure}[htbp]
\centering
\includegraphics[width=0.48\textwidth]{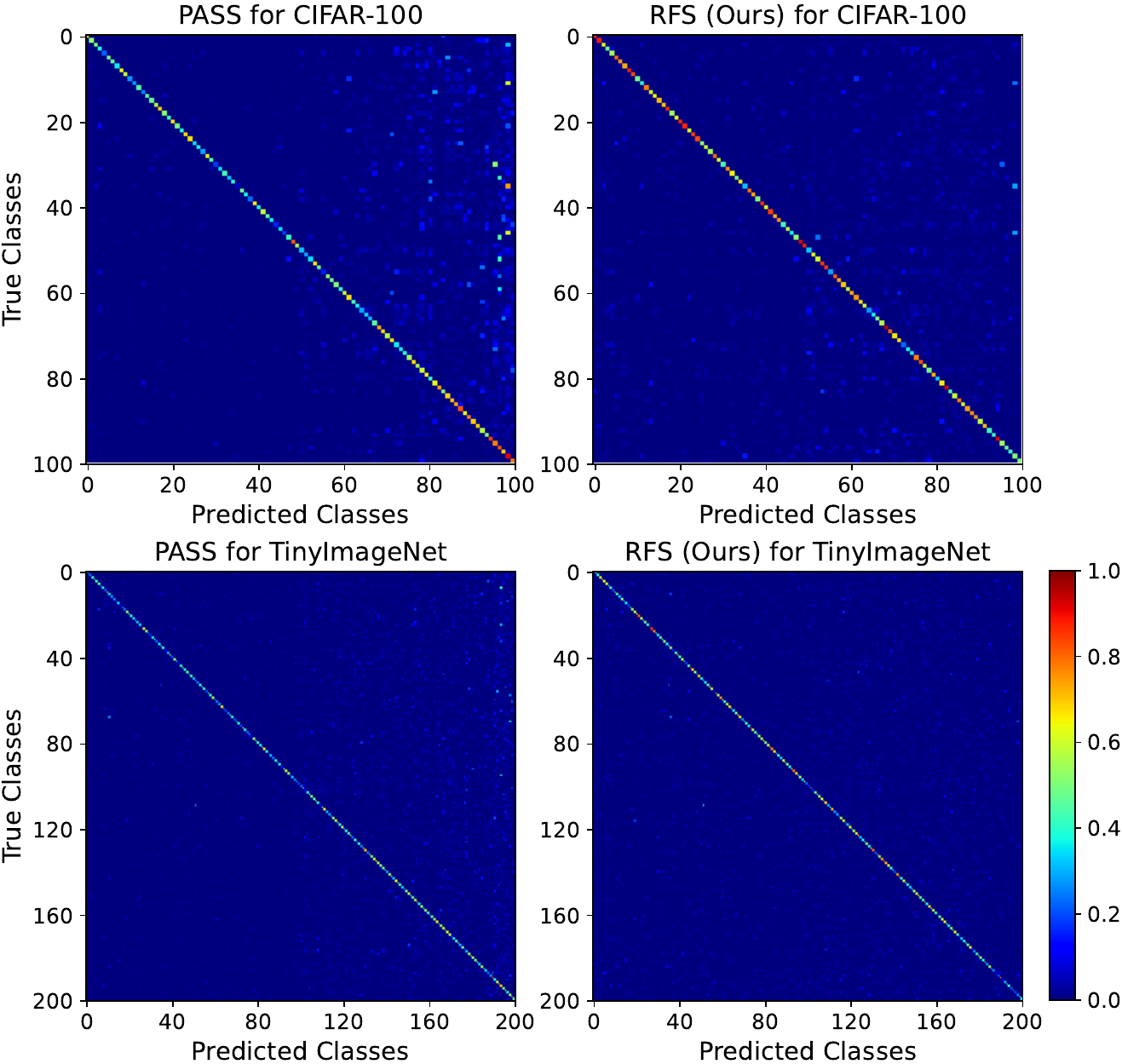}
\caption{\small Confusion matrices of PASS and our method on CIFAR-100 and TinyImageNet.}
\label{fig:confusion_matrix}
\end{figure}

\begin{figure}[htbp]
\centering
\includegraphics[width=0.48\textwidth]{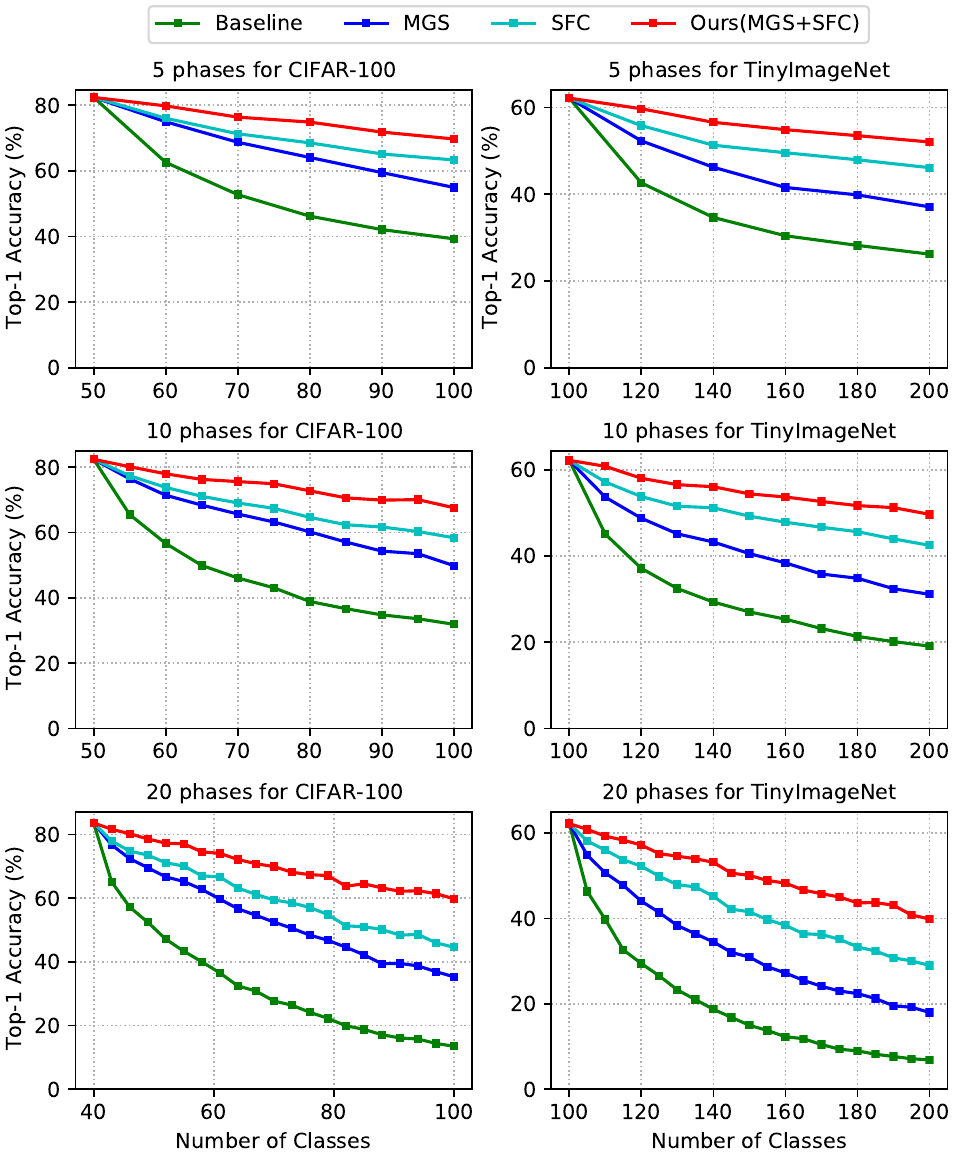}
\caption{\small Average classification accuracy of the initial classes in subsequent incremental phases on CIFAR-100 and TinyImageNet.}
\label{fig:init_acc_change}
\end{figure}

\begin{figure}[tbp]
\centering
\captionsetup[subfigure]{font=footnotesize, labelfont={footnotesize}} 
\subfloat[\scriptsize Different class orders]{\includegraphics[width=0.23\textwidth]{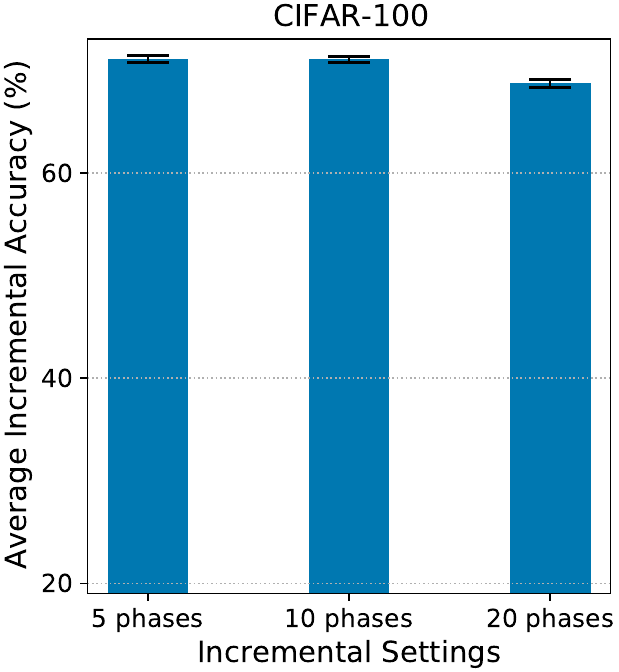}%
\label{fig:change_order_cifar}}
\hfil
\subfloat[\scriptsize Different random seeds]{\includegraphics[width=0.23\textwidth]{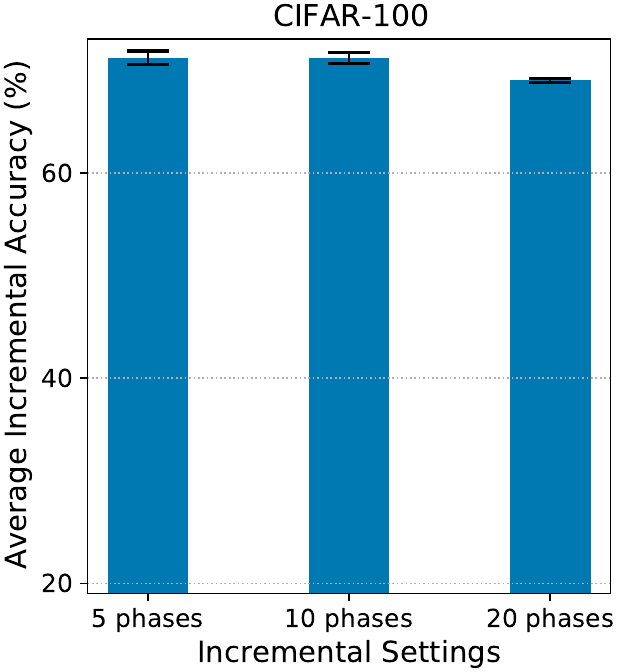}%
\label{fig:change_seed_cifar}}
\caption{\small 
Ablation study for class order and random seed. The bar chart is the middle value and the error line shows the upper and lower bound.}\vspace{-12pt}
\label{fig:rand_order_seed}
\end{figure}

\noindent \textbf{Confusion matrices analysis.} Figure \ref{fig:confusion_matrix} illustrates the confusion matrices for PASS and our proposed method in the 10 phases setting on CIFAR-100 and TinyImageNet. The diagonal entries correspond to correct predictions, while the off-diagonal entries indicate misclassifications. In the first row, PASS displays higher classification errors in the upper right triangle region. In contrast, our method shows fewer errors in the same region. A closer examination of the diagonal entries between the two matrices reveals that our method achieves higher classification accuracy than PASS across the majority of classes. Similarly, in the second row, the confusion matrices further demonstrate that our method surpasses PASS by exhibiting better classification accuracy on the diagonal entries and fewer misclassifications in the off-diagonal entries. These findings highlight the superiority of our method from a fine-grained perspective.

\noindent \textbf{Mitigating catastrophic forgetting.} The MGS and SFC components work together to synthesize retrospective features of previously learned classes, effectively mitigating catastrophic forgetting in non-exemplar class-incremental learning. Figure \ref{fig:init_acc_change} presents the average classification accuracy curve of classes learned during the initial phase as they undergo subsequent incremental phases. It is clear that the use of either MGS or SFC individually helps alleviate the significant accuracy drops of the initial classes in the later phases. Furthermore, the combined use of MGS and SFC in our method results in a more pronounced improvement in the accuracy of these initial classes. These findings highlight the significant advantages of our approach in preserving the knowledge of old classes. This improvement can be attributed to the multivariate Gaussian sampling strategy, which provides diverse and accurate representations for old classes, and the similarity-based feature compensation mechanism, which minimizes the deviation between the generated old class features and the updated classifier. Together, these two components synthesize more effective retrospective features, aiding in the reconstruction of the classification decision boundaries.

\noindent \textbf{Impact of class order and random seed.}
We investigate the impact of different class orders and random seeds on model performance. Figure \ref{fig:change_order_cifar} illustrates our method's performance across various class orders on CIFAR-100, where we rearrange the class order using three different random seeds. The median, upper, and lower bounds in the Figure indicate that our method shows slight performance fluctuations across different class orders, demonstrating low sensitivity to these variations. Additionally, Figure \ref{fig:change_seed_cifar} presents our method's performance under different random seeds. We adjust the randomness of the model training process using three different seeds while keeping the class order consistent. The results show that our method is slightly sensitive to variations in random seeds. Furthermore, as the number of incremental phases increases, the model's sensitivity further decreases. This reduced sensitivity can be attributed to the fact that our method efficiently synthesizes retrospective features through a series of operations, making it more robust to resist catastrophic forgetting.

\begin{figure}[htbp]
\centering
\includegraphics[width=0.48\textwidth]{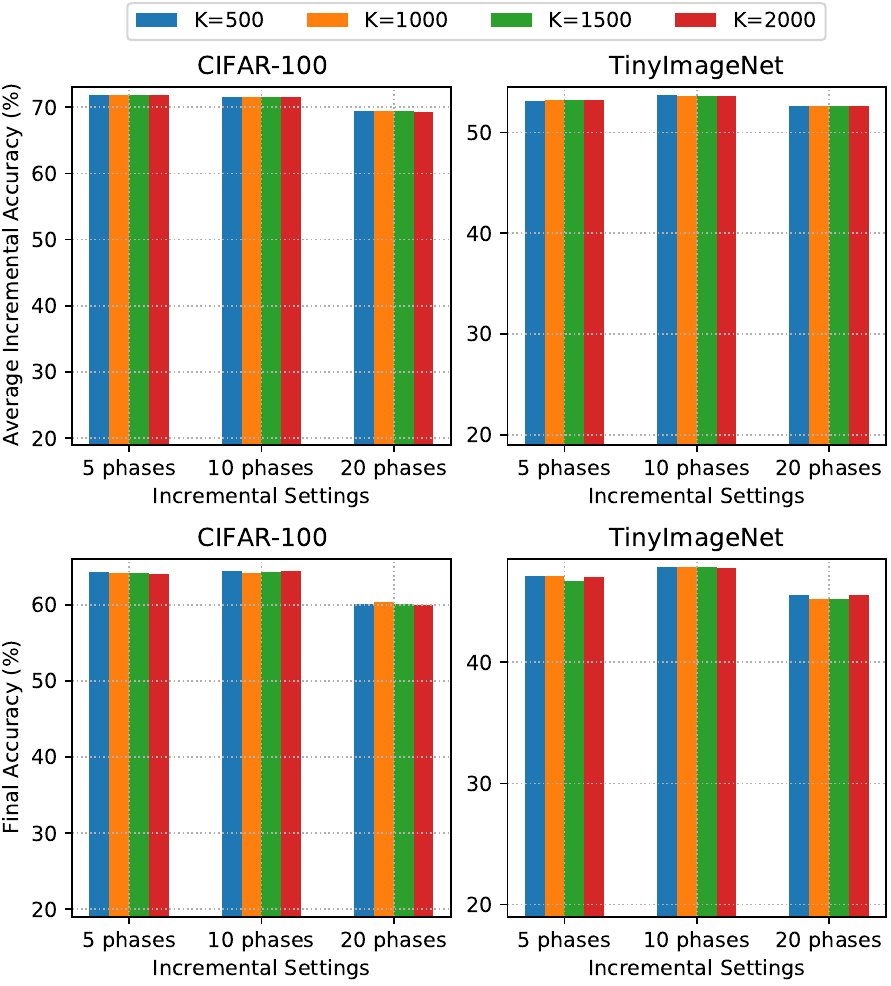}
\caption{\small Average incremental accuracy and final accuracy of our method under different K on CIFAR-100 and TinyImageNet.}
\label{fig:diff_k}
\end{figure}

\noindent \textbf{Effect of K on incremental learning performance.} Different values of K simulate various high-likelihood regions for feature sampling in MGS. To evaluate the impact of K on the incremental learning performance of our model, we compared the average incremental accuracy and final classification accuracy under different K values, as shown in Figure \ref{fig:diff_k}. The results indicate that when K is set to 500, 1000, 1500, and 2000, our method achieves almost consistent performance across different incremental settings and datasets. These results demonstrate the robustness of our method under a range of K values.

\section{Discussion}
\noindent \textbf{Storage memory and model inference analysis.} 
During training, our method does not store any raw samples of old classes. Instead, it retains only the mean and covariance matrix of the high-dimensional features for each class. While this approach slightly increases the storage memory compared to methods like PASS and PRAKA, which save only one prototype (class mean) per class, the increase is slight. This memory increase can be particularly negligible when compared to methods like iCaRL, which stores raw pixel-level samples directly. Additionally, our method applies self-supervised label augmentation only to newly learned classes during training, consuming fewer resources, such as GPU memory, than methods like PASS and PRAKA, which use label augmentation for all classes.

During inference, as outlined in Section \ref{sec:method_1}, our method only retains the feature extractor and the standard classifier. This allows it to perform on par with non-incremental methods that learn all classes in a single session, resulting in the smallest model size and fastest inference speed. In contrast, methods like PASS utilize classifiers based on label augmentation through rotation transformations, significantly increasing resource consumption during inference.

\noindent \textbf{Impact of knowledge distillation.} Our method performs knowledge distillation at both the feature level (Equation \ref{eq:feat_dist}) and the logit level (Equation \ref{eq:logit_dist}). Since knowledge distillation at the feature level is commonly used to prevent drastic changes of the feature extractor during learning new data \cite{ref_pass,ref_praka,ref_rrfe}, here we focus on the impact of logit distillation. Table \ref{tab:logit_dist} reports the performance of our method with and without logit distillation. With logit distillation, our method achieves average precision improvements of 0.14\%, 0.08\%, and 0.2\% in the three incremental settings of CIFAR-100, and 0.11\%, 0.10\%, and 0.27\% on TinyImageNet. Additionally, the method with logit distillation also improves final classification accuracy compared to the version without logit distillation. These gains can be attributed to logit distillation’s ability to enhance model stability by reducing drastic updates biased toward new classes, thereby improving overall NECIL performance.

\section{Conclusion}
In this paper, we propose a novel non-exemplar class-incremental learning method that synthesizes retrospective features for previously learned classes effectively. Our approach utilizes multivariate Gaussian distributions to model the feature space of old classes and employs a feature generation strategy that samples from high-likelihood regions to reconstruct the decision boundaries of previously learned classes. Additionally, we introduce a feature compensation mechanism that integrates the most similar new class features with generated old class representations, enhancing the model's ability to distinguish between old and new classes. Experimental results demonstrate that our method significantly improves the efficiency of non-exemplar class-incremental learning, achieving state-of-the-art performance. We believe this work provides valuable insights for advancing the field of class-incremental learning.

\bibliographystyle{IEEEtran}
\bibliography{ref}

\end{document}